\pdfoutput=1
\documentclass[conference]{IEEEtran}
\usepackage[nocompress]{cite}
\usepackage{amsmath,amssymb,amsfonts}
\usepackage{algorithmic}
\usepackage{algorithm}
\usepackage{graphicx}
\usepackage{textcomp}
\usepackage{xcolor}
\usepackage{booktabs}
\usepackage{tabularx}
\usepackage{multirow}
\usepackage{hyperref}
\usepackage{subcaption}

\graphicspath{{figures/}}

\newcommand{\sm}{\textsuperscript{\textsf{S}}}   %
\newcommand{\ex}{\textsuperscript{\textsf{E}}}   %

\makeatletter
\def\abstract{\normalfont
    \if@twocolumn
      \@IEEEabskeysecsize\bfseries\textit{\abstractname}:\ \relax
    \else
      \bgroup\par\addvspace{0.5\baselineskip}\centering\vspace{-1.78ex}\@IEEEabskeysecsize\textbf{\abstractname}\par\addvspace{0.5\baselineskip}\egroup\quotation\@IEEEabskeysecsize
    \fi\@IEEEgobbleleadPARNLSP}
\def\IEEEkeywords{\normalfont
    \if@twocolumn
      \@IEEEabskeysecsize\bfseries\textit{\IEEEkeywordsname}:\ \relax
    \else
      \bgroup\par\addvspace{0.5\baselineskip}\centering\@IEEEabskeysecsize\textbf{\IEEEkeywordsname}\par\addvspace{0.5\baselineskip}\egroup\quotation\@IEEEabskeysecsize
    \fi\@IEEEgobbleleadPARNLSP}
\makeatother

\begin{document}

\title{More GPUs or a Smaller Cache?\\ Tensor Parallelism versus KV Compression\\
for Memory-Bound LLM Serving}

\author{
\IEEEauthorblockN{Srikanta Datta Tumkur, Mehar Simhadri, Anshu Bansal, Jay Iyer,\\ Sai Pavan Kumar, Sai Kapil Kumar, Ramesh Nampelly, Raj Dandekar}
\IEEEauthorblockA{MIT and Vizuara}
}

\maketitle

\begin{abstract}
When an LLM serving deployment runs out of KV-cache room, there are two
well-established ways out, and they come from communities that rarely talk to
each other. The systems community adds GPUs. Tensor parallelism shards the
weights and the KV cache across two, four, or eight devices, buying memory
headroom at the price of an all-reduce on every layer and a hardware bill that
grows with the device count. The algorithms community shrinks the cache in
place, with KV quantisation and eviction keeping a single GPU and spending
a little quality instead. Compression papers report memory ratios, parallel-scaling
papers report throughput curves, and almost nobody puts the two on the same
cost axis, so a practitioner facing a fixed model, quality floor, and latency
target cannot tell which escape is cheaper. This paper builds that comparison.
We place tensor-parallel configurations (degree 1 to 8) and KV-compressed
configurations (16/8/4-bit, keep-ratios down to 0.25) on one cost-normalised
axis, cost per million tokens against latency, using a profiled simulator
calibrated on real A100, A40, and H100 hardware, and we go looking for the
cost-equivalence crossover. \emph{We do not find one.} Across two models
(Llama-2 at 7B and 70B), three GPU types, and every matched level of memory
relief we could construct, compression is cheaper by $1.20\times$ to
$2.00\times$, and the gap widens as relief deepens. The deeper finding is
that the question's premise fails at the scale it is usually asked. A 7B model
on an 80\,GB device cannot exhaust its KV budget within its own context window,
and the boundary that decides between the strategies is model size
relative to device memory, at roughly 36B parameters for an 80\,GB card. Below
that wall, compression dominates and extra GPUs are largely wasted spend; above
it, tensor parallelism stops being a choice and becomes an entry ticket:
Llama-2-70B is infeasible on one A100 at \emph{any} KV setting, because the
binding resource is weights, which KV compression does not touch. The two
strategies also turn out to buy different things. Tensor parallelism is the
only lever that improves latency (compression makes
per-token latency \emph{worse}, by 8 to 93\%, through batching contention), while
compression is the only lever that multiplies capacity per dollar
($16.5\times$, against $1.21\times$ for an eightfold spend on GPUs). We close
with the decision rule this evidence supports, which is simpler and more
useful than the crossover we set out to draw.
\end{abstract}


\section{Introduction}

\begin{figure*}[t]
\centering
\includegraphics[width=0.98\textwidth]{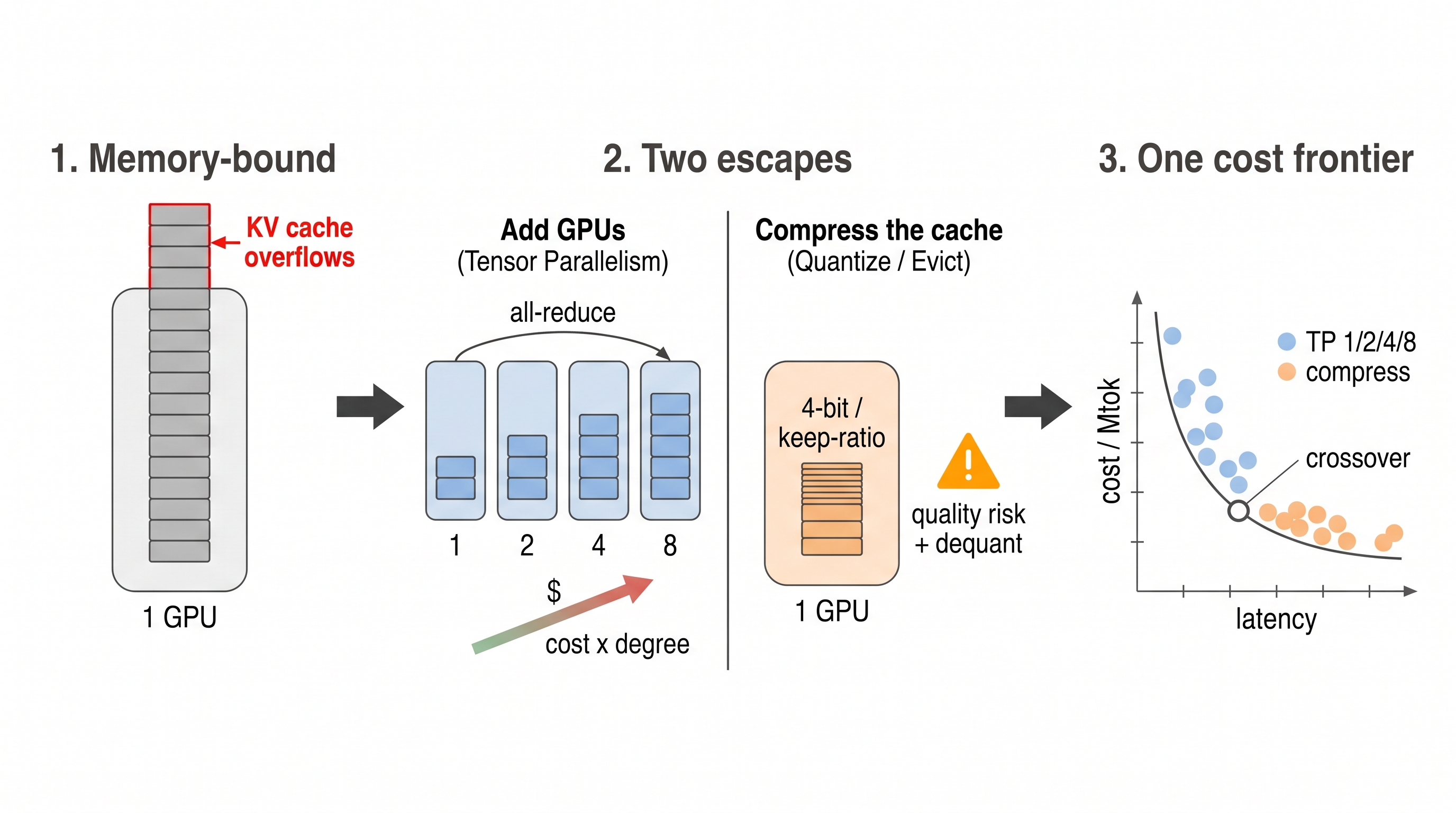}
\caption{Study overview. (1)~A memory-bound serving deployment can add GPUs
(tensor parallelism, which splits weight and KV memory but adds all-reduce
communication and multiplies cost) or compress the KV cache (quantisation and
eviction, which keep one GPU but spend quality and dequantisation overhead).
(2)~The two are reported in separate literatures, on different axes, so a
practitioner cannot tell which is cheaper. (3)~We place both on one
cost-normalised frontier (cost per million tokens against latency) at a fixed
model, quality floor, and latency target, and search for the cost-equivalence
crossover.}
\label{fig:overview}
\end{figure*}

When an LLM serving deployment becomes memory-bound, whether because the
context is long, the batch is large, or both, the KV cache stops fitting in one
GPU and something has to give. There are two established escapes, and they come from
two different communities. On the systems side, tensor
parallelism \cite{shoeybi2019megatron} shards the weights and the KV cache
across several GPUs, and context-parallel and long-context serving stacks
\cite{liu2023ringattention,kwon2023vllm} push that further, so the memory per
device falls roughly with the number of devices. On the algorithms side,
low-bit KV quantisation \cite{liu2024kivi,hooper2024kvquant}
and KV eviction \cite{zhang2023h2o,fu2025evicpress} keep a single GPU and trade
a little quality and a little dequantisation overhead for a smaller footprint
(Fig.~\ref{fig:overview}).

The problem is that these two escapes are almost never priced against each
other. Tensor-parallelism papers report a throughput or scaling curve as the
device count grows \cite{shoeybi2019megatron,zhong2024distserve};
KV-compression papers report a memory-reduction ratio at an accuracy target
\cite{liu2024kivi,hooper2024kvquant,sukhbaatar2025kvtc}; recent surveys
catalogue compression methods but stop short of a direct normalised comparison
against parallel scaling \cite{li2025rethinkkv}. Adding GPUs and compressing
the cache both buy KV headroom, but one multiplies the dollar cost by the
device count while the other costs almost nothing in hardware, so the only
fair comparison is on a \emph{cost-normalised} axis. That comparison is rarely
made, and it is the comparison an infrastructure owner needs.

This makes a concrete decision hard. An engineer facing a fixed model, a fixed
quality floor, and a fixed latency target, who must serve a longer context or
a bigger batch than one GPU holds, cannot read the two literatures and tell
whether it is cheaper to rent more GPUs or to compress the cache. The central
question, stated plainly, is: \textbf{at a fixed model, quality floor, and
latency target, is it cheaper per million tokens to add GPUs (tensor
parallelism) or to compress the KV cache, and where does the answer flip?}

We set out to build the curve that answers it. We held the model, the quality
floor, and the latency target fixed, swept tensor-parallel degree (1, 2, 4, 8)
on one side and KV-compression setting (bit-width and keep-ratio) on the
other, and plotted both as cost per million tokens against latency, expecting
to read the crossover directly off the figure.

What we found instead is the subject of this paper. The crossover does
not exist at 7B, at 70B, or on any of the three GPU types we tested, and the
premise behind the question does not survive the scale at which practitioners
ask it. We report those findings rather than
tuning the experiments until the expected curve appears, because a negative
result that survives three hardware platforms and two model scales is more
useful to a practitioner than a manufactured crossover would have been.

\subsection{Contributions}
\begin{enumerate}
\item A \textbf{cost-normalised frontier} that places tensor-parallel and
KV-compressed configurations on one axis (Fig.~\ref{fig:frontier}), on which
the uncompressed single-GPU baseline is \emph{dominated} and no crossover
appears, at either model scale or on any tested GPU
(Sec.~\ref{sec:rq1}, Sec.~\ref{sec:70b}).
\item A \textbf{decomposition of why}: tensor parallelism buys concurrency
capacity at close to a constant price per dollar while compression buys it
almost free (Fig.~\ref{fig:capdollar}), together with a sensitivity analysis
quantifying how large a dequantisation penalty would have to be to
reverse the conclusion (Fig.~\ref{fig:sens}).
\item The identification of the \textbf{real decision boundary}: model size
relative to device memory (not context length or batch size), located
between 34B and 70B parameters for an 80\,GB device across seven profiled
models (Fig.~\ref{fig:regimes}), with the resulting practitioner rule
(Table~\ref{tab:rule}).
\end{enumerate}

\section{Background and Related Work}

\subsection{Tensor parallelism and parallel scaling}
Tensor parallelism \cite{shoeybi2019megatron} splits each layer's weights, and
the attention KV cache, across several GPUs, so the per-device memory falls
roughly with the device count at the price of an all-reduce on every layer. It
is the standard way to fit a model or a long-context cache that exceeds one
GPU, and it underpins production serving
\cite{kwon2023vllm,zhong2024distserve,zheng2024sglang}. Context parallelism
pushes the same idea along the sequence dimension. Ring Attention
\cite{liu2023ringattention} distributes a long sequence across devices and
overlaps key-value communication with attention compute. Disaggregated
architectures go further still, splitting prefill from decode across machine
pools \cite{patel2024splitwise,qin2024mooncake,zhong2024distserve}. All of
these buy memory headroom by spending hardware and communication, which is one
half of our trade-off.

\subsection{KV-cache compression and eviction}
The other half keeps a single GPU and shrinks the cache. KV quantisation
stores keys and values at low bit-width. KIVI \cite{liu2024kivi} reaches 2-bit
asymmetric per-channel quantisation, and KVQuant \cite{hooper2024kvquant}
targets very long context by quantising pre-RoPE keys with non-uniform
datatypes. Eviction drops low-value tokens: H2O \cite{zhang2023h2o} keeps a
heavy-hitter set, PyramidKV \cite{cai2024pyramidkv} varies the budget by
layer, and EVICPRESS \cite{fu2025evicpress} jointly optimises compression and
eviction across storage tiers. Transform coding \cite{sukhbaatar2025kvtc}
borrows PCA and entropy coding from media compression to reach high ratios.
Each shrinks memory for a little quality and a little dequantisation overhead,
and none requires extra GPUs.

KV compression should be separated from \emph{weight} quantisation
\cite{frantar2023gptq,lin2024awq}, which shrinks the other large resident
tensor. The distinction turns out to matter a great deal in our results, because the
feasibility wall we locate in Sec.~\ref{sec:rq2} is set by weights, which KV
compression cannot touch but weight quantisation could move. Speculative
decoding \cite{leviathan2023spec,xia2024specbench} and efficient attention
kernels \cite{dao2022flashattention} accelerate inference orthogonally to both
strategies and are outside our scope.

\subsection{Why the two are rarely compared}
Parallel-scaling work reports throughput or scaling efficiency against the
device count; compression work reports a memory ratio at an accuracy target. A
recent survey \cite{li2025rethinkkv} rethinks how KV-cache compression is
measured and shows that throughput and end-to-end latency are often not
reported alongside the memory ratio, which is why a direct
normalised comparison against parallel scaling is missing. Our contribution is
to put both on the one axis a budget owner reads, and then to take
seriously what that axis shows.

\subsection{Cost-normalised comparison and Pareto dominance}
On a single cost axis, a configuration is \emph{dominated} if another is at
least as good on every axis (quality, latency, cost per million tokens) and
strictly better on one. The \emph{Pareto frontier} is the set of non-dominated
configurations; the cost-equivalence crossover, if it exists, is the point on
it where a tensor-parallel configuration and a KV-compressed configuration
meet at the same cost per million tokens. Cost per million tokens is derived
from throughput and the per-GPU hourly price, multiplied by the device count
for tensor-parallel configurations, so adding GPUs and compressing the cache
are finally on the same scale. This discipline, in which every number is
reported with its configuration stated, in the spirit of standardised inference
benchmarks \cite{reddi2020mlperf}, is as much the contribution as the numbers
themselves.

\subsection{Simulation for large configuration spaces}
The grid (parallel degree by compression setting by GPU by workload) is
combinatorial, so exhaustive measurement is infeasible. Profiled simulators
such as Vidur \cite{agrawal2024vidur} model the performance of LLM operators
from experimental profiling on real hardware and estimate end-to-end inference
latency within a reported 9\%, searching hundreds of deployment configurations
cheaply; LLMServingSim \cite{llmservingsim} couples an execution-graph
simulator to a network simulator for similar ends. We adopt Vidur as our
runtime backbone. Sec.~\ref{sec:setup} states what this does and
does not license us to claim. We take some care over that distinction, because
we had no GPU access of our own.

\subsection{Our position}
We treated the cost-equivalence curve as the deliverable: one model family,
one quality floor, one latency target, tensor-parallel and KV-compressed
configurations placed on a single cost-normalised frontier, and the crossover
reported as a function of context length and batch size. The evidence returned
something different, a feasibility wall and a pair of complementary mechanisms
instead of a crossover, and the paper reports that. No simulator
we surveyed models KV quantisation or eviction (all assume fp16 KV), and no
compression paper reports cost per million tokens, so the layer joining the
two literatures is itself a contribution, and its assumptions are our
principal exposure; we bound them explicitly in Sec.~\ref{sec:rq1}.

\section{Methodology}

\begin{figure*}[t]
\centering
\includegraphics[width=0.96\textwidth]{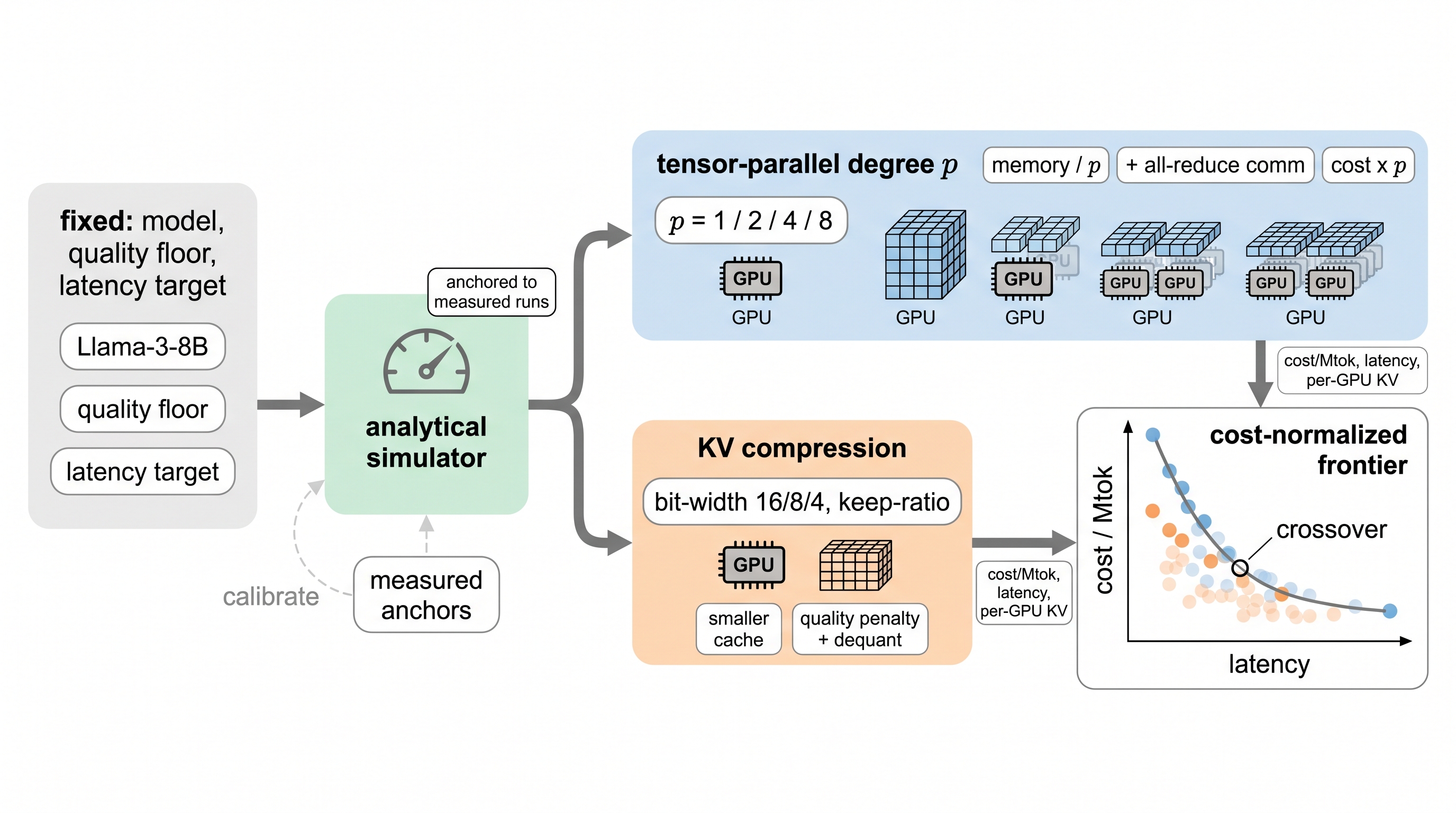}
\caption{The cost-equivalence pipeline. A fixed model, quality floor, and
latency target are held constant. On one side the tensor-parallel degree is
swept (1, 2, 4, 8): each step divides per-GPU weight and KV memory, adds
all-reduce communication, and multiplies cost by the device count. On the
other side the KV-compression setting is swept (bit-width and keep-ratio).
Both branches are evaluated through a profiled simulator and placed on a
single cost-normalised frontier (cost per million tokens against latency),
from which the dominant choice per regime, and the cost-equivalence crossover
if one exists, are read. The schematic shows the pipeline as
originally designed; the realised study substitutes Llama-2 models for the
pictured Llama-3-8B (Sec.~\ref{sec:setup}) and inherits its calibration from
the profiled backbone rather than from anchors of our own; the ``measured
anchors'' and ``crossover'' the drawing depicts were design intent, and
Sec.~\ref{sec:results} reports that no crossover was found.}
\label{fig:method}
\end{figure*}

\subsection{Axes and the cost-normalised frontier}
Every configuration is placed on three kinds of axes: \emph{latency} (TTFT and
TPOT, each at P50 and P99), \emph{throughput and memory} (tokens per second,
per-device KV occupancy, feasibility), and \emph{cost} (dollars per million
tokens). Cost is the axis that makes the two strategies comparable:
\begin{equation}
\label{eq:cost}
\mathrm{cost/Mtok}(p) = \frac{c_{\mathrm{gpu/hr}} \cdot p}{3600 \cdot
\mathrm{throughput}} \times 10^{6},
\end{equation}
where $p$ is the tensor-parallel degree; the deployment rents $p$ GPUs.
Omitting $p$ makes scale-out look free, and is in our experience the most
common way this comparison goes wrong. Dropping $p$ would reverse every
conclusion in this paper.

One axis from our original design is deliberately absent. \textbf{Quality is
not evaluated in this study.} We did not run accuracy benchmarks, and we
declined to substitute an invented degradation curve; Sec.~\ref{sec:limits}
spells out what this costs us and why it is the first thing follow-up work
should fix.

\subsection{The two strategies as one knob set}
A configuration fixes the model and chooses, on the tensor-parallel side, a
degree $p \in \{1,2,4,8\}$, and on the compression side, a KV bit-width $b \in
\{16,8,4\}$ and a keep-ratio $r_{\mathrm{keep}} \in \{1.0, 0.5, 0.25\}$.
Tensor-parallel configurations spend hardware and communication to buy memory;
KV-compressed configurations spend quality and dequantisation overhead to buy
memory. Both are evaluated under the same workloads and the same cost model,
so the only thing that separates them on the figure is what each
delivers.

\subsection{Modelling tensor parallelism}
Increasing the tensor-parallel degree $p$ has three documented effects.
(i)~Per-GPU weight and KV memory are divided by $p$ (with KV heads flooring at
the model's head count under grouped-query attention), which is what buys the
headroom. (ii)~An all-reduce communication term is added on top of the
per-token compute. (iii)~The dollar cost is multiplied by $p$. Our original
scaffold expressed the communication term analytically, as
\begin{equation}
\label{eq:tp}
t_{\text{token}}(p) = t_{\text{compute}} + \alpha\,\frac{p-1}{p}\,t_{\text{base}},
\end{equation}
with a fitted coefficient $\alpha$ that is zero at $p{=}1$ and rises and
saturates with $p$, matching the shape of a ring all-reduce. We keep the
equation for exposition but abandoned the fitted coefficient in the final
study, for a reason the results make clear. The backbone's all-reduce term is
instead driven by 6{,}958 measured collective-communication timings across
2/4/8/16 workers, and that measured behaviour produces \emph{regime-dependent}
scaling (31.9\% efficiency at TP$=8$ in one setting and a superlinear 169\% in
another, Sec.~\ref{sec:70b}) that no single $\alpha$ can express. The small
methodological lesson we take from it is to measure the communication term
instead of fitting it.

\subsection{Modelling KV compression}
Per-device KV memory is exact and definitional:
\begin{equation}
\label{eq:kv}
m_{\mathrm{kv}} = 2 \cdot L \cdot h_{\mathrm{kv}}(p) \cdot d_{\mathrm{head}}
\cdot \frac{b}{8} \cdot n_{\mathrm{ctx}} \cdot r_{\mathrm{keep}},
\end{equation}
with $L$ layers, $h_{\mathrm{kv}}(p)$ KV heads per worker after sharding, $b$
the bit-width, and $r_{\mathrm{keep}}$ the eviction keep-ratio. We implement
it by replacing the fixed two-bytes-per-element assumption in the backbone's
memory planner with $b/8$ and scaling resident tokens by $r_{\mathrm{keep}}$;
at the defaults ($b{=}16$, $r_{\mathrm{keep}}{=}1$) the patched simulator
reproduces the upstream one bit for bit.

We deliberately do \emph{not} model compression's kernel-level latency
consequences. Low-bit KV moves fewer bytes per decode step, which helps; it
also costs dequantisation work, which hurts; and we have no measurement of our
own to fix the balance. Rather than invent a curve, we let the compression arm
represent compression's \emph{best case} on latency, and then invert the
question in Sec.~\ref{sec:rq1}: how large would the penalty have to be to
change the conclusion? This converts an unmeasurable assumption into a stated,
checkable bound.

\subsection{Feasibility}
A configuration is feasible when weights and KV cache both fit:
\begin{equation}
\label{eq:feas}
\underbrace{\frac{2W}{p}}_{\text{weights}} +
\underbrace{B \cdot m_{\mathrm{kv}}}_{\text{KV for batch } B}
\;\le\; M_{\mathrm{dev}}(1-\mu),
\end{equation}
with $W$ parameters, $M_{\mathrm{dev}}$ device memory, and $\mu{=}0.1$ the
margin. Eq.~\eqref{eq:feas} carries no error term, being pure arithmetic, and
results derived from it are tagged \ex{} (exact) throughout, as distinct from
\sm{} (simulated) results that flow through the profiled backbone. Infeasible
configurations are recorded as infeasible, never silently dropped and never
costed.

\begin{algorithm}[t]
\caption{Building the frontier and searching for the crossover}
\label{alg:atlas}
\begin{algorithmic}
\STATE \textbf{Input:} model, GPUs, TP ladder, compression ladder, workloads
\STATE \textbf{for each} configuration \textbf{do}
\STATE \quad check feasibility by Eq.~\eqref{eq:feas}; record OOM explicitly
\STATE \quad \textbf{if} feasible: simulate via the profiled backbone; tag \sm{}/\ex{}
\STATE compute the Pareto frontier over (latency, cost/Mtok), feasible points only
\STATE pair TP and compressed configs at \emph{matched memory relief}
\STATE search for a sign change in $\mathrm{cost}(\mathrm{TP}) - \mathrm{cost}(\mathrm{compress})$
\STATE \textbf{return} frontier, per-regime winner, crossover \emph{or its absence}
\end{algorithmic}
\end{algorithm}

\subsection{Cost accounting}
Cost per million tokens depends on the per-GPU hourly price, the simulated
throughput, and, for tensor-parallel configurations, the device-count
multiplier $p$; all three are stated with every number. Every metric row in
the released dataset carries a provenance flag (\texttt{simulated} or
\texttt{exact}), and every figure caption in this paper repeats it. We never
present a simulated latency as measured, and we never present an assumed price
as contracted.

\section{Experimental Setup}
\label{sec:setup}

\subsection{What is measured, and by whom}
We had no GPU access. Latency and throughput come from Vidur
\cite{agrawal2024vidur}, whose operator-level predictors are fitted to
profiling collected on real A100, A40, and H100 hardware \emph{by its
authors}, with end-to-end error reported below 9\%. In our runs those
predictors fitted their profiling data at 0.29 to 0.78\% mean absolute
percentage error, and the all-reduce term is driven by measured collectives
rather than a coefficient of ours. We therefore claim no more than this:
\emph{our latency and throughput figures come from a simulator calibrated by
its authors
against real hardware}. We do not claim measured anchors of our own, and we
report no held-out simulator error, because producing one would require the
GPU access we lack. The distinction is carried through every table and figure
via the \sm{}/\ex{} tags.

\subsection{Model, hardware, and workloads}
The primary model is Llama-2-7B on an A100-80GB, with the hardware ablation
repeating the main workload on A40 and H100, and the scale ablation repeating
both workloads on Llama-2-70B. Llama-2-7B uses multi-head attention (32 KV
heads, $d_{\mathrm{head}}{=}128$, 32 layers), giving 512\,KiB of KV per token,
four times the footprint of an equivalently sized grouped-query model. That
makes it the most memory-stressing 7B choice available and biases the study
\emph{toward} finding a memory bound, not away from one. The serving
stack is Sarathi-style chunked scheduling \cite{agrawal2024sarathi} with block
size 16 \cite{kwon2023vllm} and a batch cap of 128. Principal settings are
gathered in Table~\ref{tab:hparams}.

Our original design named public workload suites for each deployment regime
(Table~\ref{tab:data}). The realised study substitutes synthetic fixed-length
workloads: \textbf{W1}, a moderate interactive load (2048 prefill, 256
decode, 64 requests, unsaturated), and \textbf{W2}, a prefill-heavy load under
saturation (3840 prefill, 256 decode, 384 requests at 30\,QPS). We chose fixed
shapes because they make the memory arithmetic exact and the runs reproducible
to the token. The mapping, and the one regime we could not reach, are recorded in the
table rather than glossed over.

\begin{table}[t]
\caption{Deployment regimes as designed, and how each was realised.}
\label{tab:data}
\centering
\footnotesize
\setlength{\tabcolsep}{4pt}
\begin{tabularx}{\columnwidth}{@{}l>{\raggedright\arraybackslash}X>{\raggedright\arraybackslash}X@{}}
\toprule
Regime & Intended source & Realised here \\
\midrule
Short chat & ShareGPT & W1 (synthetic, 2048+256, unsaturated) \\
Long-context & LongBench / RULER & \textbf{not reachable} (profiling caps at 4k) \\
Large batch & ShareGPT batched & W2 (synthetic, 3840+256, 30\,QPS) \\
Long + batched & RULER batched & W2 on Llama-2-70B (hardest bound) \\
\bottomrule
\end{tabularx}
\end{table}

\begin{table}[t]
\caption{Principal settings.}
\label{tab:hparams}
\centering
\footnotesize
\setlength{\tabcolsep}{4pt}
\begin{tabularx}{\columnwidth}{@{}l>{\raggedright\arraybackslash}X@{}}
\toprule
Setting & Value \\
\midrule
Models & Llama-2-7B (primary); Llama-2-70B (above threshold) \\
GPUs & A100-80GB, A40-48GB, H100 \\
Prices (assumed) & \$2.00, \$1.00, \$3.50 per GPU-hour \\
TP degree & 1, 2, 4, 8 (memory $/p$, measured all-reduce, cost $\times p$) \\
KV compression & bit-width 16/8/4 $\times$ keep-ratio 1.0/0.5/0.25 \\
Workloads & W1: 2048+256, 64 req; W2: 3840+256, 384 req, 30\,QPS \\
Scheduler & Sarathi-style, block size 16, batch cap 128 \\
Method & profiled simulator + exact memory model, \sm{}/\ex{} tagged \\
\bottomrule
\end{tabularx}
\end{table}

\subsection{Coverage constraint}
The backbone ships attention profiling for a fixed model set, and this
constraint shaped the study more than we expected. Llama-3-8B, our original
target, is profiled only at TP $\in \{1,4\}$ on a single GPU type, which
cannot support a tensor-parallel ladder, so Llama-2-7B (all four degrees,
three GPU types) is used instead. Context is bounded at 4096 tokens by the
model's position embeddings and the profiling range. The A40 profiling set
contains no all-reduce measurements at all, so TP${>}1$ on A40 cannot be
simulated; where that matters we say so explicitly rather than interpolating.

\section{Results}
\label{sec:results}

All numbers in this section are reproducible from
\texttt{results/metrics\_all.csv} (107 rows, each tagged \texttt{simulated} or
\texttt{exact}) via \texttt{src/make\_results.py}.

\subsection{RQ1: which is cheaper?}
\label{sec:rq1}

\begin{figure}[t]
\centering
\includegraphics[width=\columnwidth]{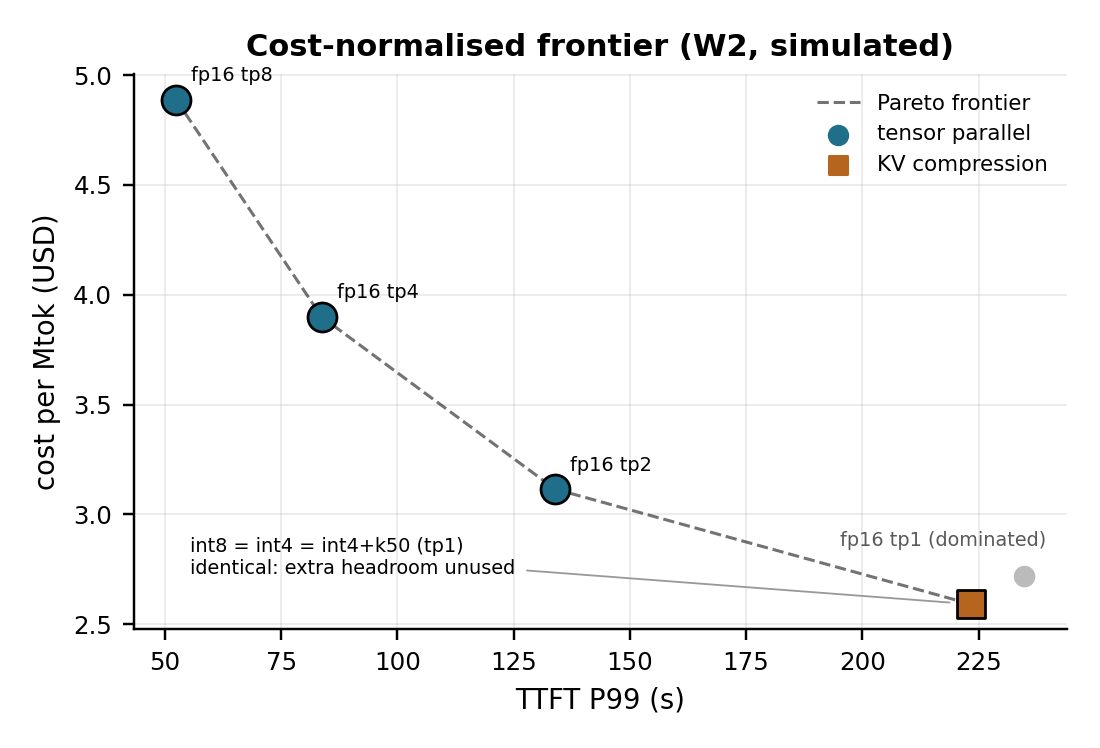}
\caption{Cost-normalised frontier on W2\sm{}. Tensor-parallel configurations
(circles) and KV-compressed configurations (squares) on cost per million
tokens against TTFT P99. The uncompressed single-GPU baseline is
\emph{dominated}: INT8 is both cheaper and faster. The three compressed points
coincide exactly, because past INT8 the additional headroom is unused in this
prefill-bound workload.}
\label{fig:frontier}
\end{figure}

\begin{table}[t]
\caption{W2 metric panel\sm{}. Llama-2-7B, A100-80GB, 3840+256 tokens,
30\,QPS. fp16/TP$=1$ is the baseline; the three compressed rows are identical
because past INT8 the memory constraint has stopped binding.}
\label{tab:main}
\centering
\footnotesize
\setlength{\tabcolsep}{3.4pt}
\begin{tabular}{@{}l r r r r r@{}}
\toprule
Config & TTFT$_{99}$ & TPOT$_{50}$ & Thr. & \$/h & \$/Mtok \\
       & (s) & (ms) & (tok/s) & & \\
\cmidrule(r){1-1}\cmidrule(lr){2-4}\cmidrule(l){5-6}
fp16 TP1 & 234.9 & 72.79 & 204.5 & 2  & 2.717 \\
fp16 TP2 & 133.9 & 48.85 & 356.7 & 4  & 3.115 \\
fp16 TP4 &  83.7 & 32.15 & 569.9 & 8  & 3.899 \\
fp16 TP8 &  52.4 & 21.58 & 909.2 & 16 & 4.888 \\
\addlinespace[2pt]
\cmidrule(r){1-1}\cmidrule(lr){2-4}\cmidrule(l){5-6}
INT8 TP1     & 223.3 & 78.83 & 214.4 & 2 & \textbf{2.591} \\
INT4 TP1     & 223.3 & 78.83 & 214.4 & 2 & \textbf{2.591} \\
INT4+k50 TP1 & 223.3 & 78.83 & 214.4 & 2 & \textbf{2.591} \\
\bottomrule
\end{tabular}
\end{table}

\begin{figure}[t]
\centering
\includegraphics[width=\columnwidth]{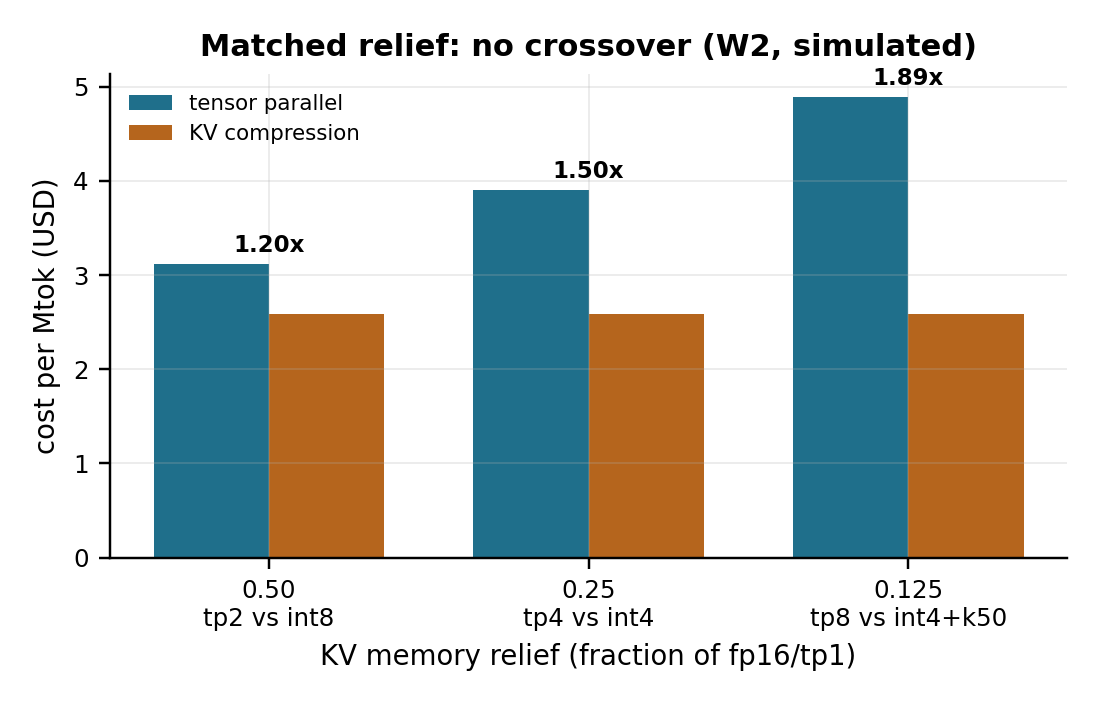}
\caption{Cost at matched KV relief\sm{}. Compression is cheaper at every level
and the ratio grows with relief: there is no cost-equivalence crossover.}
\label{fig:relief}
\end{figure}

Table~\ref{tab:main} gives the panel, and the fairest way to read it is at
\emph{matched memory relief}: the same KV bytes per device, reached by two
different means. On that comparison compression is cheaper at every level
(Fig.~\ref{fig:relief}): $1.20\times$ at half relief (TP2 vs INT8),
$1.50\times$ at quarter relief (TP4 vs INT4), and $1.89\times$ at eighth
relief (TP8 vs INT4+k50). The margin \emph{widens} as relief deepens, so no
crossover exists within the swept range, and extrapolating the two
diverging curves does not produce one.

\begin{figure}[t]
\centering
\includegraphics[width=\columnwidth]{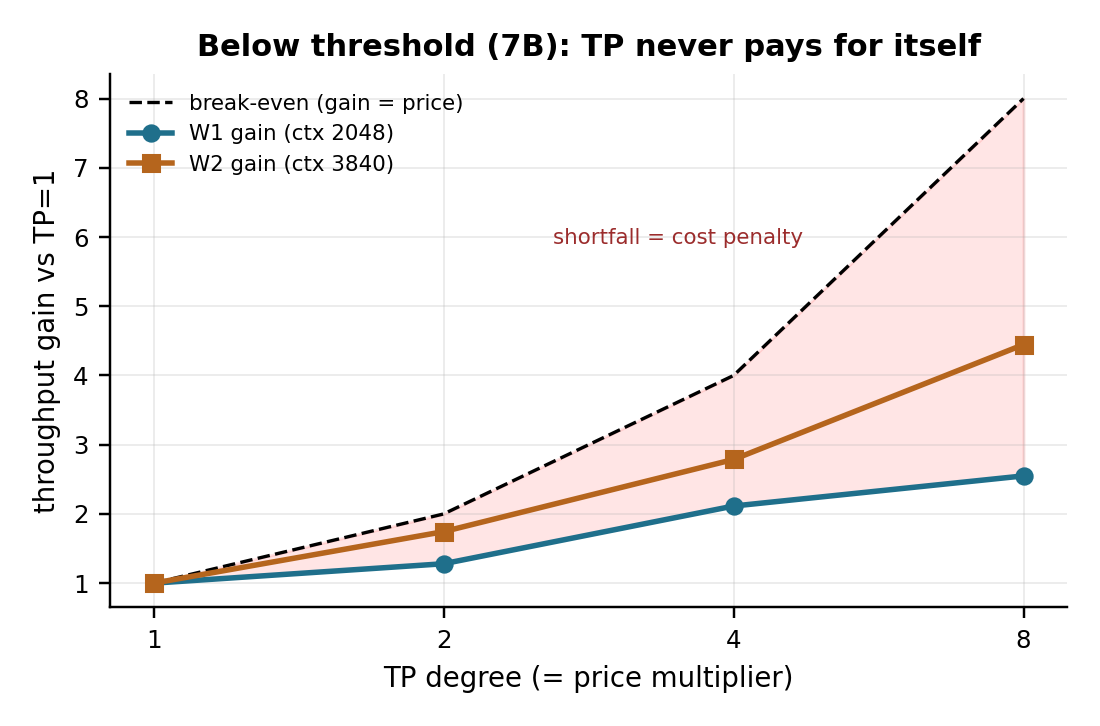}
\caption{Throughput gain against device-count price on Llama-2-7B\sm{}. The
dashed line is break-even. Both workloads sit below it at every degree, so
cost per token rises monotonically with tensor-parallel degree \emph{below the
feasibility threshold}. Sec.~\ref{sec:70b} shows this does not hold above it.}
\label{fig:scaling}
\end{figure}

Tensor parallelism does deliver on latency, cutting TTFT P99 by
$4.5\times$ from TP1 to TP8. It loses on cost because its throughput gain
never overtakes its price. On W1 the gain at TP$=8$ is $2.55\times$ for
$8\times$ the spend (31.9\% scaling efficiency); on the prefill-heavy W2 it
is $4.45\times$
(55.6\%). Both sit below break-even at every degree (Fig.~\ref{fig:scaling}),
so each added GPU raises the cost of every token it serves.

\begin{figure}[t]
\centering
\includegraphics[width=\columnwidth]{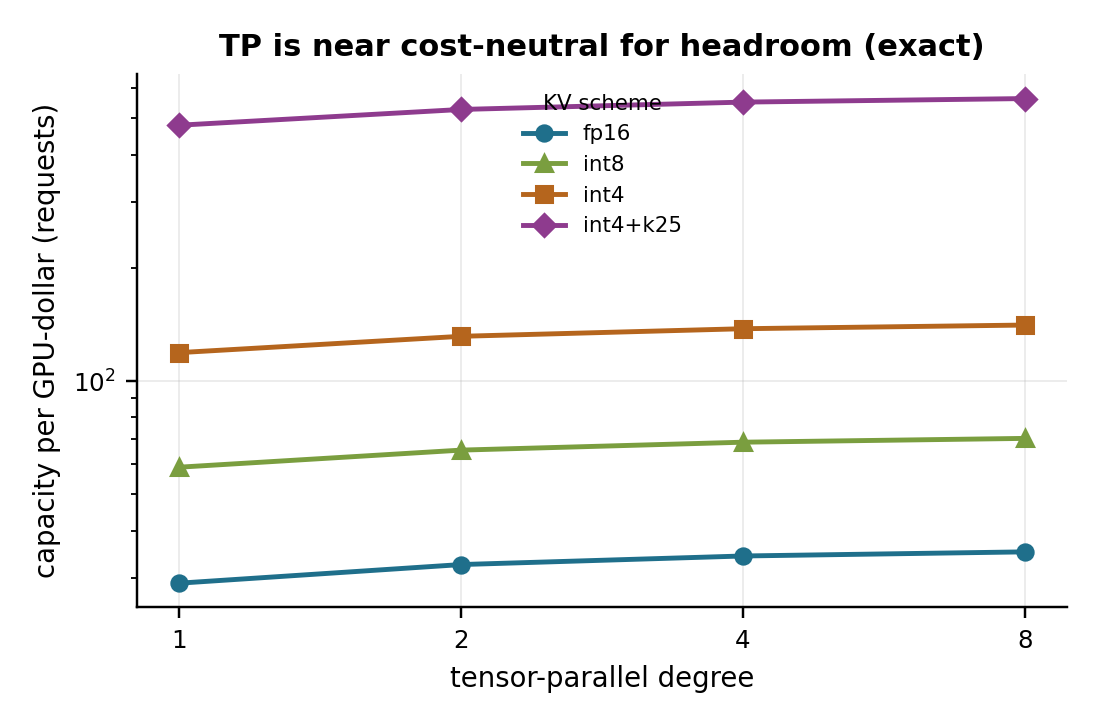}
\caption{Concurrency capacity per GPU-dollar\ex{}, ctx 4096. Moving right
along any curve (more GPUs) is nearly flat; moving between curves (more
compression) is a step change.}
\label{fig:capdollar}
\end{figure}

The same fact looks starker as capacity per unit spend
(Fig.~\ref{fig:capdollar}\ex{}). An eightfold increase in GPU spend improves
concurrency capacity per GPU-dollar from $29.0$ to just $35.1$ ($+21\%$, the
modest gain coming from weight sharding); as a way to buy headroom, tensor
parallelism is very nearly \emph{cost-neutral}. Compression lifts the same
figure from $29.0$ to $479.0$, a $16.5\times$ gain, at unchanged hardware
cost. This asymmetry, not any subtlety of scheduling, is the mechanism behind
the absent crossover.

\textbf{Compression's latency cost.} The saving comes with a penalty, and this
is where the two strategies part ways most sharply. Compression
\emph{degrades} per-token latency. At 7B, INT4 raises TPOT P50 from 72.79 to
78.83\,ms ($+8\%$, Table~\ref{tab:main}); at 70B, from 116.76 to 224.81\,ms
($+93\%$, Table~\ref{tab:70b}). The mechanism is batching; our compression
arm models no dequantisation cost at all. The freed memory admits more
concurrent sequences, and the added contention lengthens every decode step.
Compression therefore trades TPOT for capacity, while tensor parallelism is
the only lever that improves TTFT and TPOT simultaneously.
Under a tight per-token latency target, scale-out is the only direction that
helps.

\begin{figure}[t]
\centering
\includegraphics[width=\columnwidth]{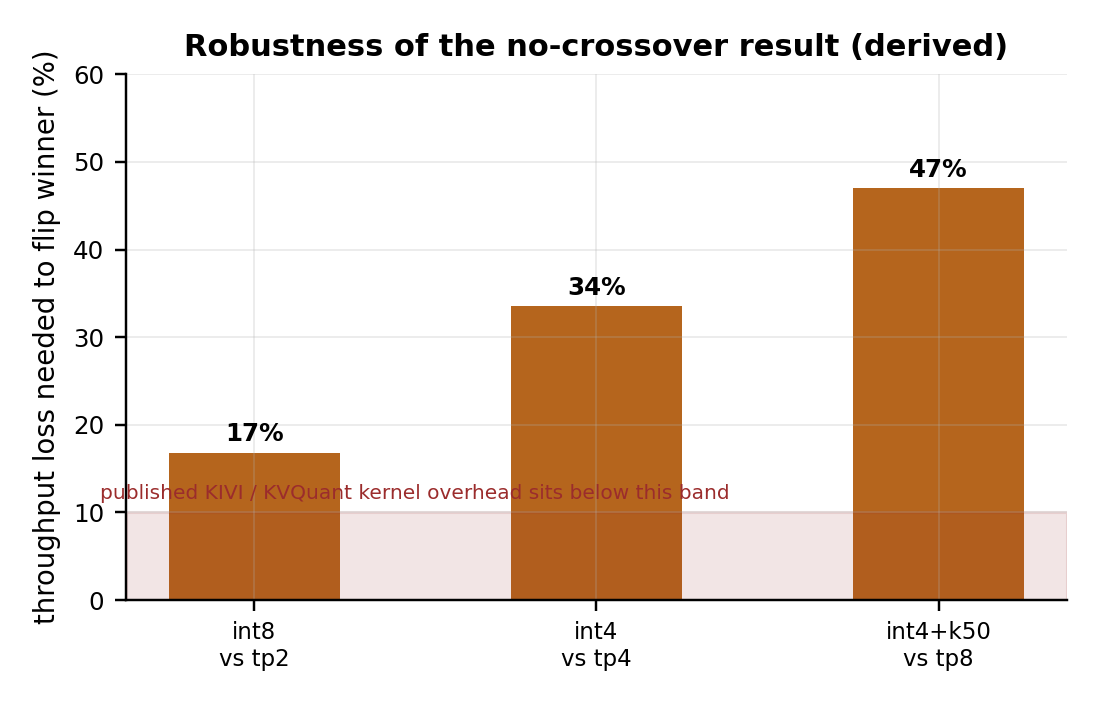}
\caption{Sensitivity of the result to unmodelled dequantisation overhead. Bars
give the throughput loss compression would need to suffer before tensor
parallelism becomes cheaper.}
\label{fig:sens}
\end{figure}

\textbf{Robustness.} Because our compression arm carries no dequantisation
penalty, one might reasonably worry that the comparison is rigged in its
favour. So we invert the question and ask how much throughput compression
would have to lose before tensor parallelism wins. The answer is 16.8\%, 33.5\%, and
47.0\% at the three relief levels (Fig.~\ref{fig:sens}). Published low-bit KV
kernels \cite{liu2024kivi,hooper2024kvquant} report overheads well below the
smallest of these, so the ordering is robust to the one assumption we could
not measure, though confirming it on hardware remains the most valuable
experiment this study leaves undone.

\subsection{RQ2: where is the crossover, and how does it move?}
\label{sec:rq2}

RQ2 asks how the crossover moves as context length and batch size grow. We
report two findings, the first of which invalidates the question's premise at
typical deployment scale.

\textbf{A 7B model on an 80\,GB device cannot be driven out of memory.}\ex{}
Sweeping reserved tokens per request, fp16/TP$=1$ retains capacity for 29
concurrent requests at the model's maximum 4096-token context, and reaches OOM
(inability to admit even one request) only at 131{,}072 tokens,
$32\times$ beyond the context window. Within its own limits, this
configuration cannot exhaust its KV budget. The frontier agrees. The
compressed points in Fig.~\ref{fig:frontier} are identical to three decimal
places, because past INT8 the memory constraint has already stopped binding
and further relief is inert.

\begin{figure}[t]
\centering
\includegraphics[width=\columnwidth]{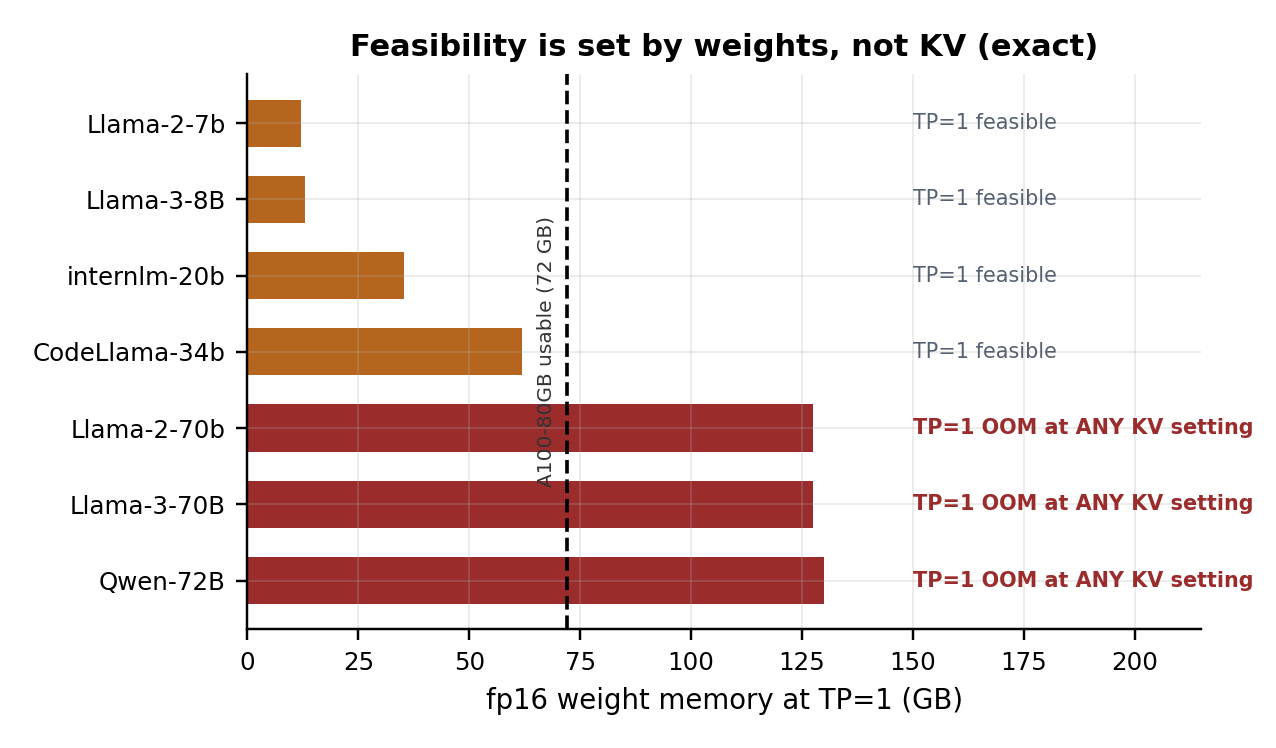}
\caption{Feasibility at TP$=1$ across the profiled model set\ex{}. Above the
device budget, no KV setting restores feasibility: the constraint is weights.}
\label{fig:regimes}
\end{figure}

\textbf{The real boundary is model size.}\ex{} Extending
Eqs.~\eqref{eq:kv} to \eqref{eq:feas} across the seven models the backbone
profiles (Fig.~\ref{fig:regimes}, Table~\ref{tab:regimes}) locates a hard
feasibility wall. For Llama-2-70B and Llama-3-70B, fp16 weights require
127.5\,GB per device at TP$=1$ against 72\,GB usable, so TP$=1$ is infeasible,
and \emph{remains infeasible under INT4 with a 0.25 keep-ratio}, because the
binding resource is weights, which KV compression does not shrink. Between
CodeLlama-34B (61.9\,GB, feasible) and Llama-2-70B (127.5\,GB, infeasible)
lies the boundary: roughly 36B parameters at fp16 on a 72\,GB budget. The
decision between our two strategies turns on which side of this wall a
deployment sits on; context length and batch size do not decide it.

\begin{table}[t]
\caption{Concurrency capacity at ctx 4096\ex{}; 0 denotes OOM.}
\label{tab:regimes}
\centering
\footnotesize
\setlength{\tabcolsep}{4pt}
\begin{tabular}{@{}l r r r r r@{}}
\toprule
Model & Wt.\ TP1 & fp16 & fp16 & INT4+k25 & INT4+k25 \\
      & (GB)     & TP1  & TP4  & TP1      & TP4 \\
\midrule
Llama-2-7b     &  12.1 & 29  & 137 & 479  & 2207 \\
Llama-3-8B     &  13.0 & 118 & 550 & 1888 & 8800 \\
internlm-20b   &  35.4 & 7   & 53  & 124  & 862  \\
CodeLlama-34b  &  61.9 & 13  & 301 & 216  & 4824 \\
Llama-2-70b    & 127.5 & \textbf{0} & 128 & \textbf{0} & 2054 \\
Llama-3-70B    & 127.5 & \textbf{0} & 128 & \textbf{0} & 5740 \\
Qwen-72B       & 130.0 & \textbf{0} & 15  & \textbf{0} & 252  \\
\bottomrule
\end{tabular}
\end{table}

\subsection{Above the threshold: Llama-2-70B}
\label{sec:70b}

Everything so far characterises the strategies \emph{outside} the regime the
premise describes, so we repeated both workloads on Llama-2-70B, which sits
above the wall, with identical settings.

The wall shows up in execution as well as in arithmetic. The TP$=1$ run
terminates in the memory planner, unable to admit a single request. Both arms
consequently begin at TP$=2$, and memory relief is measured relative to
fp16/TP$=2$. In this regime memory constrains the deployment, which at fp16/TP$=2$ runs
at 99 to 100\% KV occupancy holding roughly 13 concurrent requests. Compression correspondingly stops being inert: INT8 and
INT4 now yield distinct throughputs (68.3 versus 71.8 tok/s), where at 7B all
compressed configurations were identical. The inertness point has moved
rather than vanished, since INT4 and INT4+k25 still coincide, memory having
ceased to bind between them.

\begin{table}[t]
\caption{W2 on Llama-2-70B\sm{}. Both arms start at TP$=2$, which is the
baseline here, because TP$=1$ is infeasible.}
\label{tab:70b}
\centering
\footnotesize
\setlength{\tabcolsep}{4.2pt}
\begin{tabular}{@{}l r r r r@{}}
\toprule
Config & Thr. & KV mem & \$/h & \$/Mtok \\
       & (tok/s) & (\%) & & \\
\cmidrule(r){1-1}\cmidrule(lr){2-3}\cmidrule(l){4-5}
fp16 TP1 & \multicolumn{4}{c}{\textbf{INFEASIBLE}: 127.5\,GB weights} \\
\addlinespace[2pt]
fp16 TP2 &  53.3 & 99.1 & 4  & 20.846 \\
fp16 TP4 & 123.8 & 25.8 & 8  & 17.950 \\
fp16 TP8 & 184.5 &  9.2 & 16 & 24.089 \\
\addlinespace[2pt]
\cmidrule(r){1-1}\cmidrule(lr){2-3}\cmidrule(l){4-5}
INT8 TP2     & 68.3 & 97.8 & 4 & 16.268 \\
INT4 TP2     & 71.8 & 63.5 & 4 & \textbf{15.475} \\
INT4+k25 TP2 & 71.8 & 15.6 & 4 & \textbf{15.475} \\
\bottomrule
\end{tabular}
\end{table}

\begin{figure}[t]
\centering
\includegraphics[width=\columnwidth]{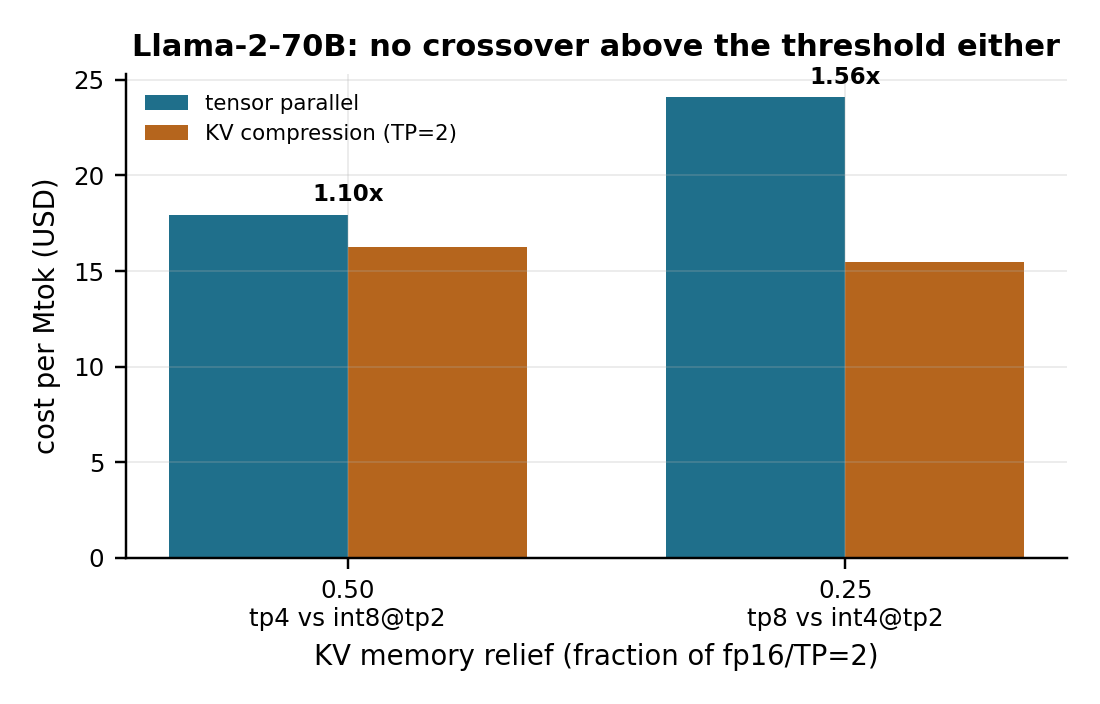}
\caption{Matched relief on Llama-2-70B\sm{}. Compression remains cheaper above
the threshold, where both arms start at TP$=2$.}
\label{fig:relief70}
\end{figure}

The crossover is still absent (Fig.~\ref{fig:relief70}). At matched relief,
compression is cheaper by $1.10\times$ (INT8/TP$=2$ at \$16.27 against
fp16/TP$=4$ at \$17.95) and by $1.56\times$ (INT4/TP$=2$ at \$15.48 against
fp16/TP$=8$ at \$24.09). That the headline result holds in both regimes
strengthens it considerably, because it is not an artefact of studying a
model that was never memory-bound.

\begin{figure}[t]
\centering
\includegraphics[width=\columnwidth]{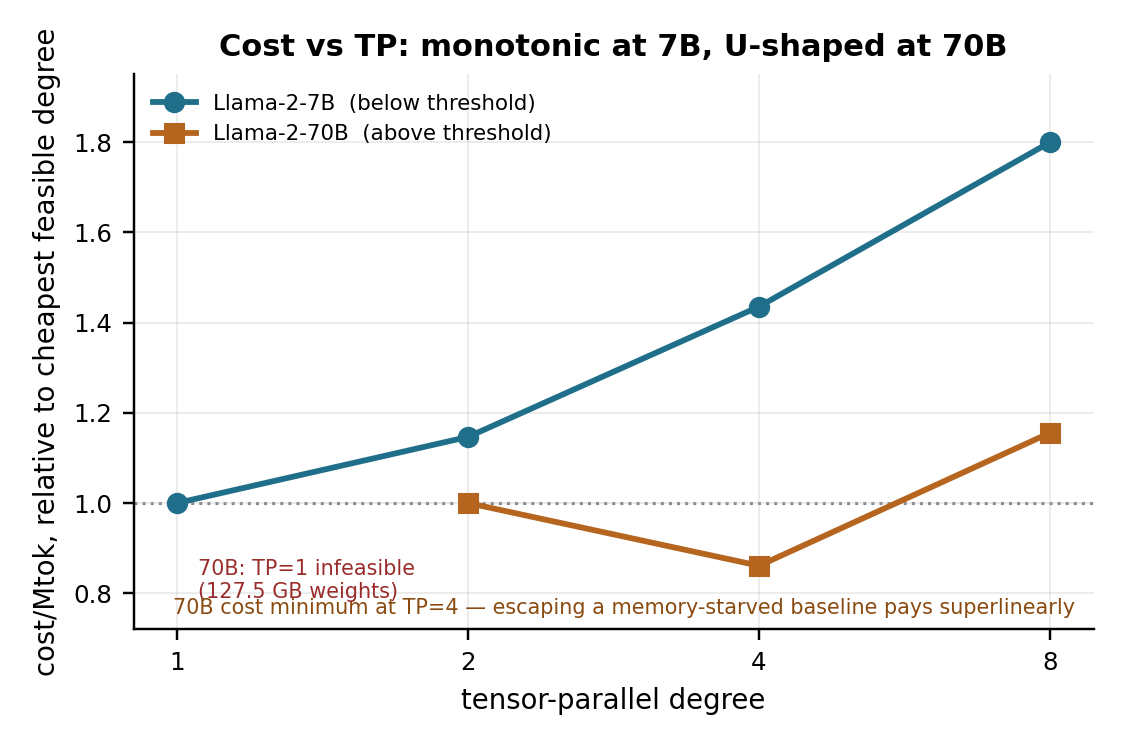}
\caption{Cost per million tokens against parallel degree, normalised to each
model's cheapest feasible degree\sm{}. Monotonically increasing below the
threshold; U-shaped above it.}
\label{fig:degree}
\end{figure}

One result does reverse, however. Below the
threshold, cost per token rises monotonically with parallel degree
(Fig.~\ref{fig:scaling}). Above it, cost is \emph{U-shaped}, with a minimum
at TP$=4$ (Fig.~\ref{fig:degree}). On W1 the step from TP$=2$ to TP$=4$
delivers $3.38\times$ the throughput for $2\times$ the price (169\%
efficiency) and reduces cost per token by 41\% (by 14\% on W2). The effect is
neither an anomaly nor free performance. At TP$=2$ the deployment is
capacity-starved rather than compute-starved, so the added devices relieve
the binding constraint \emph{and} add compute; superlinear scaling is the
signature of escaping a binding constraint, and it is a one-time payoff:
beyond TP$=4$ the effect exhausts and cost rises again.

The practical consequence is that $p_{\min}$, the minimum feasible degree, is
not necessarily the cheapest: when $p_{\min}$ runs memory-saturated, one
degree above it is cheaper per token. Sec.~\ref{sec:disc} folds this into the
decision rule.

\subsection{RQ3: how does the answer shift with hardware and degree?}

Across degrees, the all-reduce term erodes benefit as hypothesised.
Scaling efficiency falls to 31.9\% at TP$=8$ on W1, so high degrees never sit
on the cost frontier unless feasibility forces them (Fig.~\ref{fig:scaling}).

\begin{figure}[t]
\centering
\includegraphics[width=\columnwidth]{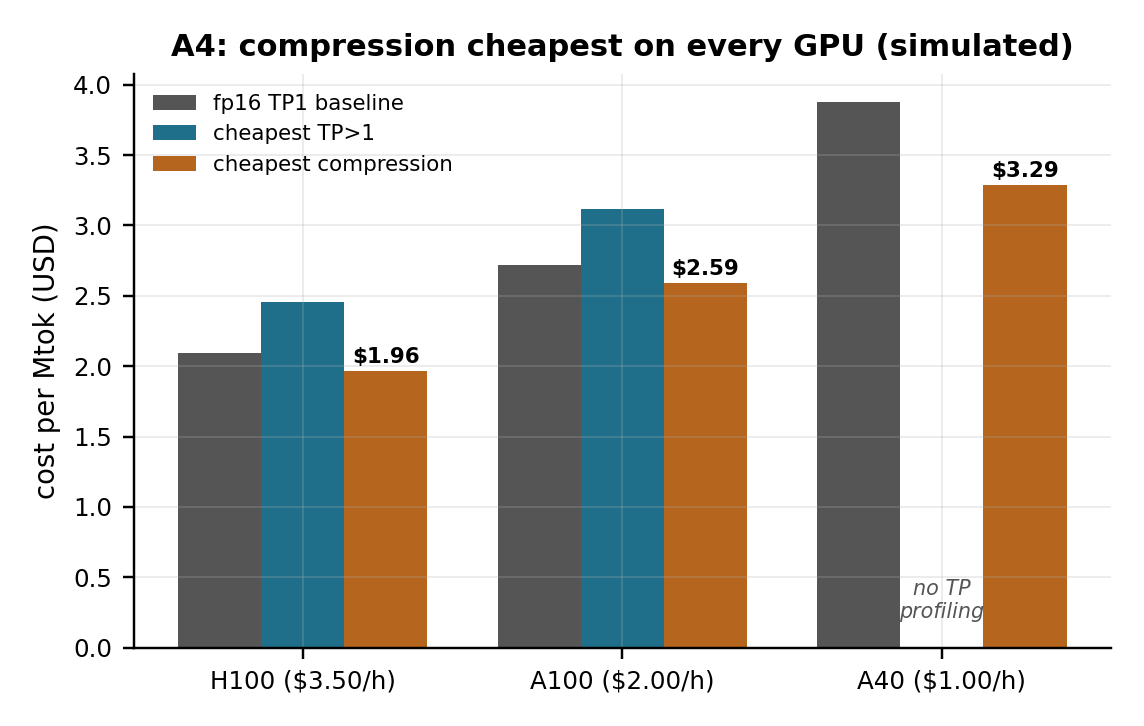}
\caption{Ablation A4\sm{}: cheapest configuration per arm across GPU types
(W2, 7B). Compression wins on every device; per-token cost orders opposite to
hourly price.}
\label{fig:hardware}
\end{figure}

Across hardware, we repeated W2 on A40 (\$1.00/h assumed) and H100 (\$3.50/h
assumed); Fig.~\ref{fig:hardware}. Three findings emerged, one of them
surprising to us. First, \textbf{the ordering is
hardware-invariant}: compression is the cheapest arm on every device, with
matched-relief margins on H100 ($1.25\times$, $1.48\times$, $2.00\times$)
mirroring A100. That is unsurprising within a device, once one notices that
the hourly price cancels in the TP-vs-compression ratio, leaving only the
throughput profile. Second, \textbf{the cheap GPU is not the cheap
deployment}: cost per token orders H100 (\$1.96) $<$ A100 (\$2.59) $<$ A40
(\$3.29), the reverse of hourly price, because the A40's throughput deficit
exceeds its price advantage. Third, the A40 (48\,GB) is memory-bound even at
7B (99\% occupancy at fp16), and it is therefore the one device where
INT4 outperforms INT8 (84.5 vs 82.7 tok/s); compression's relief stops
being inert where memory still binds, confirming the mechanism of
Sec.~\ref{sec:rq1} from the hardware side. The folk rule ``compress on cheap
GPUs'' thus survives, but for a different reason than usually given. The low
hourly price cancels, and what matters is the smaller memory that makes cheap
devices memory-bound sooner. The one coverage limit here is that the A40 profiling
set contains no all-reduce measurements, so TP${>}1$ on A40 cannot be
simulated, and its TP arm is absent rather than measured-and-losing.

\section{Discussion}
\label{sec:disc}

Taken together, the evidence does not support treating the two strategies as
substitutable escapes to be priced against one another. They relieve
different constraints, and most of this paper's surprises dissolve once that
is said plainly. Tensor parallelism is a \emph{feasibility} mechanism: it
shards weights, which is the only way past the wall in
Fig.~\ref{fig:regimes}, and as a way to buy KV headroom it is close to
cost-neutral (Fig.~\ref{fig:capdollar}). Compression is a \emph{capacity}
mechanism: it multiplies concurrency roughly $16\times$ per dollar, but it
does nothing for weights, and nothing at all once memory has stopped binding,
as the coincident points in Fig.~\ref{fig:frontier} show. Only tensor
parallelism buys \emph{latency}; compression sells it
(Sec.~\ref{sec:rq1}).

They are complementary, and the ordering matters: choose the minimum degree
that makes the weights fit, $p_{\min} = \lceil 2W / M_{\mathrm{usable}}
\rceil$, then compress for concurrency. At 70B this is visible directly in
Table~\ref{tab:regimes}, where TP$=4$ restores feasibility at 128 requests and
INT4+k25 then takes the same deployment to 2054. Sec.~\ref{sec:70b} adds one
correction, namely that $p_{\min}$ is the cheapest degree only when it is not
itself memory-saturated. For Llama-2-70B, $p_{\min}=2$ runs at 99 to 100\% occupancy,
and stepping to TP$=4$ reduces cost per token by 41\% on W1 and 14\% on W2
while more than doubling throughput. The rule, in full: take $p_{\min}$ for
feasibility; if that configuration runs memory-saturated, take one degree
more, because escaping the constraint pays superlinearly; then compress.
Beyond that point additional degrees stop paying, and compression remains the
cheaper source of headroom at every matched level in both regimes.

\begin{table}[t]
\caption{Scale-out versus compress: the decision rule.}
\label{tab:rule}
\centering
\small
\begin{tabular}{p{2.0cm}p{1.95cm}p{3.65cm}}
\toprule
Regime & Choose & Because \\
\midrule
Weights exceed device memory ($\gtrsim$36B at fp16, 80\,GB) &
TP at $p_{\min}$; compression cannot substitute &
KV compression does not shrink weights (Table~\ref{tab:regimes}) \\
\addlinespace
$p_{\min}$ runs memory-saturated &
Step up one degree &
$3.38\times$ throughput for $2\times$ price; cost falls 41\% (W1) and 14\% (W2), Sec.~\ref{sec:70b} \\
\addlinespace
Weights fit; KV is binding &
Compress first &
$16.5\times$ capacity per dollar vs $1.21\times$ (Fig.~\ref{fig:capdollar}) \\
\addlinespace
Weights fit; KV not binding &
Neither; compression is inert &
Coincident points, Fig.~\ref{fig:frontier} \\
\addlinespace
Latency target unmet &
TP, and pay for it &
$4.5\times$ TTFT P99 cut; compression moves TPOT the \emph{wrong} way
($+8$ to $93\%$) \\
\addlinespace
Quality floor forbids compression &
TP &
Only remaining source of headroom \\
\bottomrule
\end{tabular}
\end{table}

\subsection{Relation to our stated hypothesis}
We pre-registered the expectation that tensor parallelism would become
\emph{cheaper} once the cache no longer fits on one GPU. That hypothesis is
\textbf{not supported}, and we want to be explicit about how it
failed, because the failure is instructive. In the regime where both
strategies are feasible, compression is cheaper at every matched relief level
and the gap widens; in the regime where compression fails, tensor parallelism
is \emph{compulsory}, and it ``wins'' there only in the sense that there is
no contest. The corrected claim, that the switch is governed by
weight footprint rather than by context length, is a stronger practitioner
rule than the one we set out to confirm, and we report it in place of the
original.

\section{Limitations}
\label{sec:limits}

\textbf{No measured anchors of our own.} We had no GPU access. Latency and
throughput are inherited from a simulator calibrated by its authors
\cite{agrawal2024vidur}; we report no held-out error against hardware we
measured, and the fidelity ablation our study design called for could not be
performed. All \ex{}-tagged results are unaffected, being closed-form.

\textbf{Quality was not evaluated.} No accuracy benchmark was run. Every
compression result should therefore be read as an upper bound on what
compression delivers, valid only while the quality floor permits the setting.
Since the quality floor is what forbids aggressive compression,
and thus what forces tensor parallelism in practice, this is the most
consequential gap in the study and the first thing later work should close.

\textbf{Compression latency is unmodelled.} We quantify this exposure rather
than hide it. Fig.~\ref{fig:sens} gives the penalty that would reverse each
comparison.

\textbf{The wall assumes fp16 weights.} Our $\sim$36B boundary is a statement
about fp16 weight footprints. Weight quantisation
\cite{frantar2023gptq,lin2024awq} shrinks the resource that sets the
wall and would move it upward, since a 70B model at 4-bit weights fits where its
fp16 form does not. The wall's \emph{existence}, and the complementarity of
the two mechanisms, are unaffected; its numeric position should be read as
specific to fp16 serving.

\textbf{Coverage.} One model family (Llama-2 at 7B and 70B); context bounded
at 4096 tokens, so the long-context regime of Table~\ref{tab:data} is
unreached; batch cap 128 (which binds for TP4, TP8, and INT4+k25); synthetic
workloads rather than the public suites originally named. The hardware
ablation covers A100, A40, and H100 at 7B only; the 70B runs are A100-only;
A40 lacks all-reduce profiling so its TP arm cannot be simulated.

\textbf{Prices.} \$2.00 (A100), \$1.00 (A40), and \$3.50 (H100) per GPU-hour
are representative on-demand cloud rates at the time of writing, chosen for
round-number transparency; they are assumed, not contracted, and readers should
substitute their own. Eq.~\eqref{eq:cost} is linear in price, so a
uniform change rescales all costs without altering any within-device ordering;
the TP-vs-compression winner on a given device is price-invariant.
The \emph{cross}-device ordering of Fig.~\ref{fig:hardware} does depend on
relative prices and should be re-costed for a reader's own contract.

\section{Conclusion}

We set out to price tensor parallelism against KV compression on one axis and
name the crossover. There is no crossover, in either regime. Compression is
cheaper at every matched level of memory relief (by $1.20$ to $1.89\times$
below the feasibility threshold and $1.10$ to $2.00\times$ above it and across
hardware), and would need to lose 17 to 47\% of its throughput to
dequantisation before the ordering reversed. The mechanism is simple once
seen: tensor parallelism buys headroom at roughly the price of the hardware,
while compression buys it nearly free.

The more useful result concerns the premise rather than the answer. A 7B
model on an 80\,GB device cannot exhaust its KV
budget within its context window; the decision boundary is model size relative
to device memory, not context length or batch size, at roughly 36B
parameters for an 80\,GB device under fp16 weights. Above the wall, tensor
parallelism is compulsory and compression cannot substitute; below it,
compression dominates and additional GPUs are largely wasted spend. The two
regimes also differ in a way that qualifies the usual advice to minimise
device count: below the wall, every added GPU raises cost per token, but
above it, when the minimum feasible degree is itself memory-saturated, the
next degree up is cheaper, because escaping a binding constraint pays a
one-time superlinear return.

What we would do next follows from what is missing: close the quality axis,
whose floor is the real force pushing deployments toward tensor parallelism;
obtain measured anchors for low-bit KV decode kernels, the one assumption we
could only bound; and re-locate the wall under quantised weights, which is
where the practitioner questions now live.

\section*{Acknowledgments}
The authors thank Vizuara AI Labs for mentorship. This study used publicly
released profiling data from the Vidur project \cite{agrawal2024vidur}; no
GPU measurements were performed by the authors.

\section*{Reproducibility}
\texttt{src/make\_results.py} regenerates \texttt{results/metrics\_all.csv}
(107 rows, each tagged \texttt{simulated} or \texttt{exact}) and every results
figure in this paper. The compression layer is a documented modification to
the backbone's memory planner implementing Eq.~\eqref{eq:kv}, vendored as a
patch with the backbone pinned by commit; with defaults ($b{=}16$,
$r_{\mathrm{keep}}{=}1$) it reproduces upstream behaviour exactly.
\texttt{results/error\_taxonomy.md} records worked examples of the failure
modes this methodology guards against.

\end{document}